\documentclass[10pt, a4paper]{article}
\usepackage{lrec}
\usepackage{multibib}
\newcites{languageresource}{Language Resources}
\usepackage{graphicx}
\usepackage{tabularx}
\usepackage{soul}

\usepackage{epstopdf}
\usepackage[latin1]{inputenc}

\usepackage{hyperref}
\usepackage{xstring}
\usepackage{color}

\title{WorldTree: A Corpus of Explanation Graphs for Elementary Science Questions\\supporting Multi-Hop Inference}

\name{Peter A. Jansen$^{*}$, Elizabeth Wainwright$^{\dag}$, Steven Marmorstein$^{\ddag}$, Clayton T. Morrison$^{*}$}

\address{$^{*}$School of Information, $^{\dag}$Department of Linguistics, $^{\ddag}$Department of Computer Science \\
 University of Arizona, Tucson, USA \\
         pajansen@email.arizona.edu\\}

\abstract{
Developing methods of automated inference that are able to provide users with compelling human-readable justifications for why the answer to a question is correct is critical for domains such as science and medicine, where user trust and detecting costly errors are limiting factors to adoption. One of the central barriers to training question answering models on explainable inference tasks is the lack of gold explanations to serve as training data.  In this paper we present a corpus of explanations for standardized science exams, a recent challenge task for question answering.  We manually construct a corpus of detailed explanations for nearly all publicly available standardized elementary science question (approximately 1,680 $3^{rd}$ through $5^{th}$ grade questions) and represent these as ``explanation graphs'' -- sets of lexically overlapping sentences that describe how to arrive at the correct answer to a question through a combination of domain and world knowledge.  We also provide an explanation-centered tablestore, a collection of semi-structured tables that contain the knowledge to construct these elementary science explanations.  Together, these two knowledge resources map out a substantial portion of the knowledge required for  answering and explaining elementary science exams, and provide both structured and free-text training data for the explainable inference task. \\ 
\newline 
 }

\begin{document}

\maketitleabstract

\section{Introduction}

\noindent Question answering (QA) is a high-level natural language processing task that requires automatically providing answers to natural language questions.  The approaches used to construct QA solvers vary depending on the questions and domain, from inference methods that attempt to construct answers from semantic, syntactic, or logical decompositions, to retrieval methods that work to identify passages of text likely to contain the answer in large corpora using statistical methods.  Because of the difficulty of this task, overall QA task performance tends to be low, with generally between 20\% and 80\% of natural (non-artificially generated) questions answered correctly, depending on the questions, the domain, and the knowledge and inference requirements. 

\indent Standardized science exams have recently been proposed as a challenge task for question answering \cite{clark:2015}, as these questions have very challenging knowledge and inference requirements \cite{clark:2013,jansen2016:COLING}, but are expressed in simple-enough language that the linguistic challenges are likely surmountable in the near-term.  They also provide a standardized comparison of modern inference techniques against human performance, with individual QA solvers generally answering between 40\% to 50\% of multiple choice science questions correctly \cite[inter alia]{Khot2015ExploringML,Clark2016CombiningRS,Khashabi:2016TableILP,Khot:ACL2017,jansen2017framing}, and top-performing ensemble models nearly reaching a passing grade of 60\% on middle school ($8^{th}$ grade) science exams during a recent worldwide competition of 780 teams sponsored by the Allen Institute for AI \cite{schoenick2017moving}. 

One of the central shortcomings of question answering models is that while solvers are steadily increasing the proportion of questions they answer correctly, most solvers generally lack the capacity to provide human-readable explanations or justifications for why those answers are correct.  This ``explainable inference'' task is seen as a limitation of current machine learning models in general (e.g. Ribeiro et al., \shortcite{Ribeiro2016}), but is critical for domains such as science or medicine where user trust and detecting potentially costly errors are important.  More than this, evidence from the cognitive and pedagogy literature suggests that explanations (when tutoring others) and self-explanations (when engaged in self-directed learning) are an important aspect of learning, helping humans better generalize the knowledge they have learned \cite{roscoe2007understanding,legare2014contributions,rittle2016eliciting}.  This suggests that explainable methods of inference may not only be desirable for users, but may be a requirement for automated systems to have human-like generalization and inference capabilities.

Building QA solvers that generate explanations for their answers is a challenging task, requiring a number of inference capacities.  Central among these is the idea of \textit{information aggregation}, or the idea that explanations for a given question are rarely found in a contiguous passage of text, and as such inference methods must generally assemble many separate pieces of knowledge from different sources in order to arrive at a correct answer.  Previous estimates \cite{jansen2016:COLING} suggest elementary science questions require an average of 4 pieces of knowledge to answer and explain those answers (here our analysis suggests this is closer to 6), but  inference methods tend to have difficulty aggregating more than 2 pieces of knowledge from free-text together due to the semantic or contextual ``drift'' associated with this aggregation \cite{fried2015higher}.  Because of the difficulty in assembling training data for the information aggregation task, some have approached explanation generation as a distant supervision problem, with explanation quality modelled as a latent variable \cite{jansen2017framing,sharp:2017tellmewhy}.  While these techniques have had some success in constructing short explanations, semantic drift likely limits the viability of this technique for explanations requiring more than two pieces of information to be aggregated.

\begin{figure}[t]
\begin{center}
\includegraphics[scale=0.30]{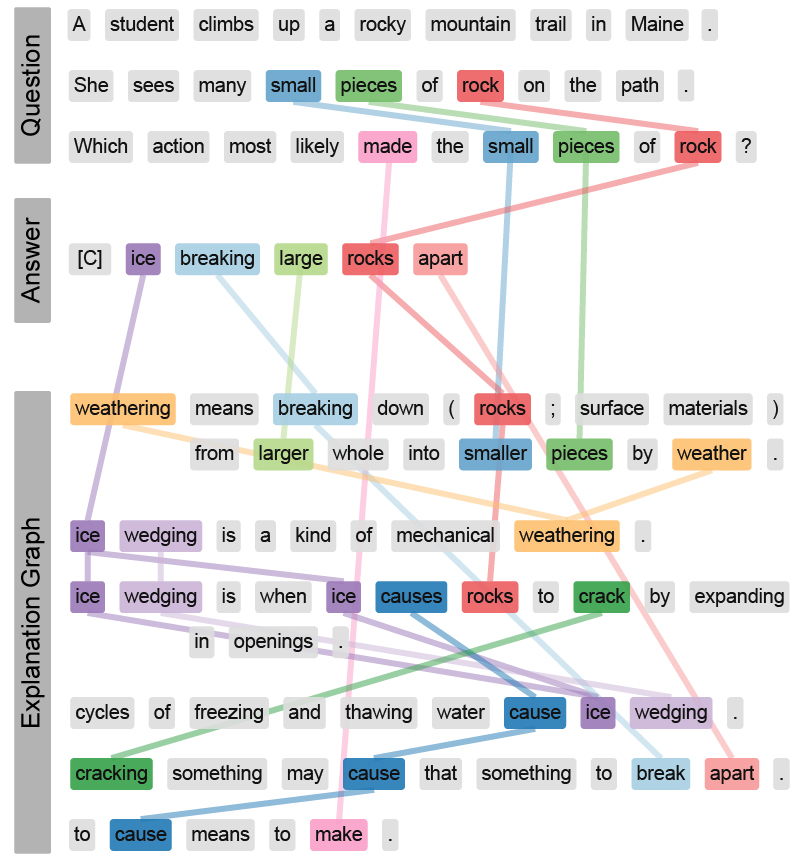} 
\caption{\small An example multiple choice science question, the correct answer, and a sample explanation graph for why that answer is correct.  Here, the explanation graph consists of six sentences, each interconnected through lexical overlap with the question, answer, and other explanation sentences.  }
\vspace{-7mm}
\label{fig:examplequestionexplanation}
\end{center}
\end{figure}

To address this, here we construct a large corpus of explanation graphs (see Figure \ref{fig:examplequestionexplanation}) to serve as training data for explainable inference tasks. 
The contributions of this work are are: 
\begin{itemize}
\item We construct a set of explanations for 1,680 standardized elementary science exam questions, represented as both free-text, and as lexically-overlapping ``explanation graphs'' that provide training data for inference models by detailing explicit connections between knowledge in different sentences of an explanation. 
\item We provide an explanation-centered ``tablestore'', a set of 62 semi-structured tables containing 4,950 rows that provide a substantial portion of the knowledge required to answer non-spatial, non-mathematical elementary science questions. 
\item We provide an analysis of the knowledge growth and explanation overlap properties of this corpus, suggesting both  requirements for inference algorithms to make use of explanation corpora, as well as methods of estimating the difficulty in constructing explanation corpora in other domains.
\end{itemize}

\section{Related Work}

\noindent In terms of question answering, the ability to provide compelling human-readable explanations for answers to questions has been proposed as a complementary metric to assess QA performance alongside the proportion of questions answered correctly.  Jansen et al. \shortcite{jansen2017framing} developed a QA system for elementary science that answers questions by building and ranking explanation graphs built from aggregating multiple sentences read from free text corpora, including study guides and dictionaries.  Because of the difficulty in constructing gold explanations to serve as training data, the explanations built with this system were constructed by modeling explanation quality as a latent variable machine learning problem. First, sentences were decomposed into sentence graphs based on clausal and prepositional boundaries, then assembled into multi-sentence ``explanation graphs''.  Questions were answered by ranking these candidate explanation graphs, using answer correctness as well as features that capture the connectivity of key-terms in the graphs as a proxy for explanation quality.  Jansen at al. \shortcite{jansen2017framing} showed that it is possible to learn to generate high quality explanations for 60\% of elementary science questions using this method, an increase of 15\% over a baseline that retrieved single continuous passages of text as answer justifications. 
Critically, in their error analysis Jansen et al. found that for questions answered incorrectly by their system, nearly half had successfully generated high-quality explanation graphs and ranked these highly, though they were not ultimately selected.  They suggest that the process of building and ranking explanations would be aided by developing more expensive second-pass reranking processes that are able to better recognize the components and structure of high-quality explanations within a short list of candidates. 

Knowledge bases of tables, or \textit{``table stores''}, have recently been proposed as a semi-structured knowledge formalism for question answering that balances the cost of manually crafting highly-structured knowledge bases with the difficulties in acquiring this knowledge from free text \cite{yin:2015answering,sun:2016table,jauhar:2016tables}. The methods for question answering over tables generally take the form of constructing chains of multiple table rows that lead from terms in the question to terms in the answer, while the tables themselves are generally either collected from the web, automatically generated by extracting relations from free text, or manually constructed. 

At the collection end of the spectrum, Pasupat and Liang \shortcite{pasupat:2015} extract 2,108 HTML tables from Wikipedia, and propose a method of answering these questions by reasoning over the tables using formal logic. They also introduce the WikiTableQuestions dataset, a set of 22,033 question-answer pairs (such as \textit{``Greece held its last Summer Olympics during which year?''}) that can be answered using these tables.  Demonstrating the ability for collection at scale, Sun et al. \shortcite{sun:2016table} extract a total of 104 million tables from Wikipedia and the web, and develop a model that constructs relational chains between table rows using a deep-learning framework.\footnote{Sun et al. \shortcite{sun:2016table} note that the 99 million tables extracted from the web introduce more noise into the inference process than the high-quality tables from Wikipedia}  Using their system and table store, Sun et al. demonstrate state-of-the-art performance on several benchmark datasets, including WebQuestions \cite{berant2013semantic}, a set of popular questions asked from the web designed to be answerable using the large structured knowledge graph Freebase (e.g. \textit{``What movies does Morgan Freeman star in?''}). 

In terms of automatic generation, though relations are often represented as \textit{$<subject, relation, argument>$} triples, Yin et al. \shortcite{yin:2015answering} create a large table containing 120M \textit{n-tuple} relations using OpenIE \cite{etzioni:2011open}, arguing that the extra expressivity afforded by these more detailed relations allows their system to answer more complex questions. Yin et al. use this to successfully reason over the WebQuestions dataset, as well as their own set of questions with more complex prepositional and adverbial constraints. 

Elementary science exams contain a variety of complex and challenging inference problems \cite{clark:2013,jansen2016:COLING}, with nearly 70\% of questions requiring some form of causal, process, or model-based reasoning to solve and produce an explanation for.  In spite of these exams being taken by millions of students each year, elementary students tend not to be fast or voluminous readers by adult standards, making this a surprisingly low-resource domain for grade-appropriate study guides and other materials.  The questions also tend to require world knowledge expressed in grade-appropriate language (like that \textit{bears have fur} and that \textit{fur keeps animals warm}) to solve.  Because of these requirements and limitations, table stores for elementary science QA tend to be manually or semi-automatically constructed, and comparatively small. 

Khashabi et al. \shortcite{Khashabi:2016TableILP} provide the largest elementary science table store to date, containing approximately 5,000 manually-authored rows across 65 tables based on science curriculum topics obtained from study guides and a small corpus of questions.  Khashabi et al. also augment their tablestore with 4 tables containing 2,600 automatically generated table rows using OpenIE triples.  Reasoning is accomplished using an integer-linear programming algorithm to chain table rows, with Khashabi et al. reporting that an average of 2 table rows are used to answer each question. Evaluation on a small set of 129 science questions achieved passing performance (61\%), with an ablation study showing that the bulk of their model's performance was from the manually authored tables.  

To help improve the quality of automatically generated tables, Dalvi et al. \shortcite{Dalvi2016IKE} introduce an interactive tool for semi-automatic table generation that allows annotators to query patterns over large corpora.  They demonstrate that this tool can improve the speed of knowledge generation by up to a factor of 4 over manual methods, while increasing the precision and utility of the tables up to seven fold compared to completely automatic methods. 

All of the above systems share the commonality that they work to connect (or aggregate) multiple pieces of knowledge that, through a variety of inference methods, move towards the goal of answering questions.  Fried et al. \shortcite{fried2015higher} report that information aggregation for QA is currently very challenging, with few methods able to combine more than two pieces of knowledge before succumbing to \textit{semantic drift}, or the phenomenon of two pieces of knowledge being erroneously connected due to shared lexical overlap, incomplete word-sense disambiguation, or other noisy signals (e.g. erroneously aggregating a sentence about \textit{Apple computers} to an inference when working to determine whether \textit{apples} are a kind of fruit).  In a generating a corpus of natural-language explanations for 432 elementary science questions, Jansen et al. \shortcite{jansen2016:COLING} found that the average question requires aggregating 4 separate pieces of knowledge to explainably answer, with some questions requiring much longer explanations.  

Though few QA solvers explicitly report the aggregation limits of their algorithms, Fried et al. \shortcite{fried2015higher}, Khabashi et al. \shortcite{Khashabi:2016TableILP} and Jansen et al. \shortcite{jansen2017framing} appear to show limits or substantial decreases in performance after aggregating two pieces of knowledge.  To the best of our knowledge, of systems that use information aggregation, only Jansen et al. \shortcite{jansen2017framing} explicitly rate the explanatory performance of the justifications from their model, with good explanations generated for only 60\% of correctly answered questions.  Taken together, all of this suggests that performance on information aggregation and explainable question answering is still far from human performance, and could substantially benefit from a large corpus of training data for these tasks.

%
%
%
%
%
%

\section{Design Goals}

\noindent We began with the following design goals:

{\flushleft \textbf{Computable explanations:}} Explanations should be represented at different levels of structure (explanation, then sentences, then relations within sentences).  The knowledge links between explanation sentences should be explicit through lexical overlap, which can be used to form an ``explanation graph'' that describes how each sentence is linked in an explanation. 

{\flushleft \textbf{Depth:}} Sufficient knowledge should be present in explanations such that that the answer could be arrived at with little extra domain or world knowledge -- i.e. where possible, explanations should be targeted at the level of knowledge of a 5-year old child, or lower (see below for a more detailed discussion of explanatory depth).

{\flushleft \textbf{Reuse:}} Where possible, knowledge should be re-used across explanations to facilitate automated analysis of knowledge use, and identifying common explanation patterns across questions. 


\subsection{Explanation Depth}

%
%
\begin{table}[t]
\footnotesize
\centering
\begin{tabular}{lll}
Question	&	\multicolumn{2}{l}{ Which of the following characteristics would }\\
			&	\multicolumn{2}{l}{ best help a tree survive the heat of a forest fire?	}\\
Answers		&	[A] large leaves			&	[B] shallow roots	\\
			&	\textit{[*C] thick bark}	&	[D]	thin trunks		\\

\hline
\multicolumn{3}{l}{ \textit{Levels of explanatory knowledge:} }\\
\\

\multicolumn{3}{l}{ \textit{Domain Expert (e.g. teacher)} }\\
\multicolumn{3}{l}{ ~~Bark is a protective covering around the trunk and branches }\\
\multicolumn{3}{l}{ ~~of a tree. }\\
\\
\multicolumn{3}{l}{ \textit{Domain Novice (e.g. 4th grade student)} }\\
\multicolumn{3}{l}{ ~~As an object's thickness increases, it's resistance to damage}\\
\multicolumn{3}{l}{ ~~will also increase. }\\
\\
\multicolumn{3}{l}{ \textit{Young child (e.g. 5-year old)} }\\
\multicolumn{3}{l}{ ~~Protecting something means preventing harm.  }\\
\multicolumn{3}{l}{ ~~Fire causes harm to trees, forests, and other living things. }\\
\multicolumn{3}{l}{ ~~Thickness is a measure of how thick an object is. }\\
\multicolumn{3}{l}{ ~~A tree is a kind of living thing. }\\
\\
\multicolumn{3}{l}{ \textit{First Principles} }\\
\multicolumn{3}{l}{ ~~Protecting a living thing has a positive impact on it's}\\
\multicolumn{3}{l}{ ~~survival and health. }\\
\hline

\end{tabular}
\caption{\small Levels of explanatory knowledge depth in order of increasing specificity, and example explanatory sentences for each level.  For a domain expert who is already fluent in the reasoning of a domain, brief explanations may be sufficient to completely understand why a given answer is correct.  As the level of explanatory knowledge moves towards increasing specificity, less domain and world knowledge is assumed, and this knowledge must be explicitly included in the explanations.  Explanatory levels are additive, i.e. an explanation targeted at the \textit{young child} level would also include the knowledge at the \textit{domain novice} and \textit{domain expert} levels.  In this work, we target authoring explanations at a level between \textit{young child} and \textit{first principles}.}
\label{tab:explanatorydepth}
\vspace{-4mm}
\end{table}

\noindent The level of knowledge required to convincingly explain why an answer to a question is correct depends upon one's familiarity with the domain of the question.  For a domain expert (such as an elementary science teacher), a convincing explanation to why \textit{thick bark} is the correct answer to \textit{''Which characteristic could best help a tree survive the heat of a forest fire?''} might need only take the form of explaining that \textit{one of bark's primary functions is to provide protection for the tree}.  In contrast, for a domain novice, such as an elementary science student, this explanation might need to be elaborated to include more knowledge to make this inference, such as that \textit{thicker things tend to provide more protection}. 
Here we identify four coarse levels of increasing explanatory knowledge depth, shown in Table \ref{tab:explanatorydepth}. 

For training explainable inference systems, a high level of explanatory depth is likely required.  As such, in this work we target authoring explanations between the levels of \textit{young child} and \textit{first principles}.  Pragmatically, in spite of their ultimate utility for training inference systems, building explanations too close to \textit{first principles} becomes laborious and challenging for annotators given the level of abstraction and the large amount of implicit world knowledge that must be enumerated, and we leave developing protocols and methods for building such detailed explanations for future work.

\section{Explanation Authoring}

\noindent We describe our representations, tools, and annotation process below. 

\subsection{Questions}
\noindent We author explanation graphs for a corpus of 2,201 elementary science questions ($3^{rd}$ through $5^{th}$ grade) from the AI2 Science Questions V2 corpus, consisting of both standardized exam questions from 12 US states, as well as the separate AI2 Science Questions Mercury dataset, a set of questions licensed from a student assessment entity.  Each question is a 4-way multiple choice question, and only those questions that do not involve diagram interpretation (a separate spatial task) are included.  Approximately 20\% of explanations required specialized domain knowledge (for example, spatial or mathematical knowledge) that did not easily lend itself to explanation using our formalism, resulting in a corpus of 1,680 questions and explanations. 

\subsection{Tables and Table Rows}
\begin{figure*}[t]
\begin{center}
\includegraphics[scale=0.40]{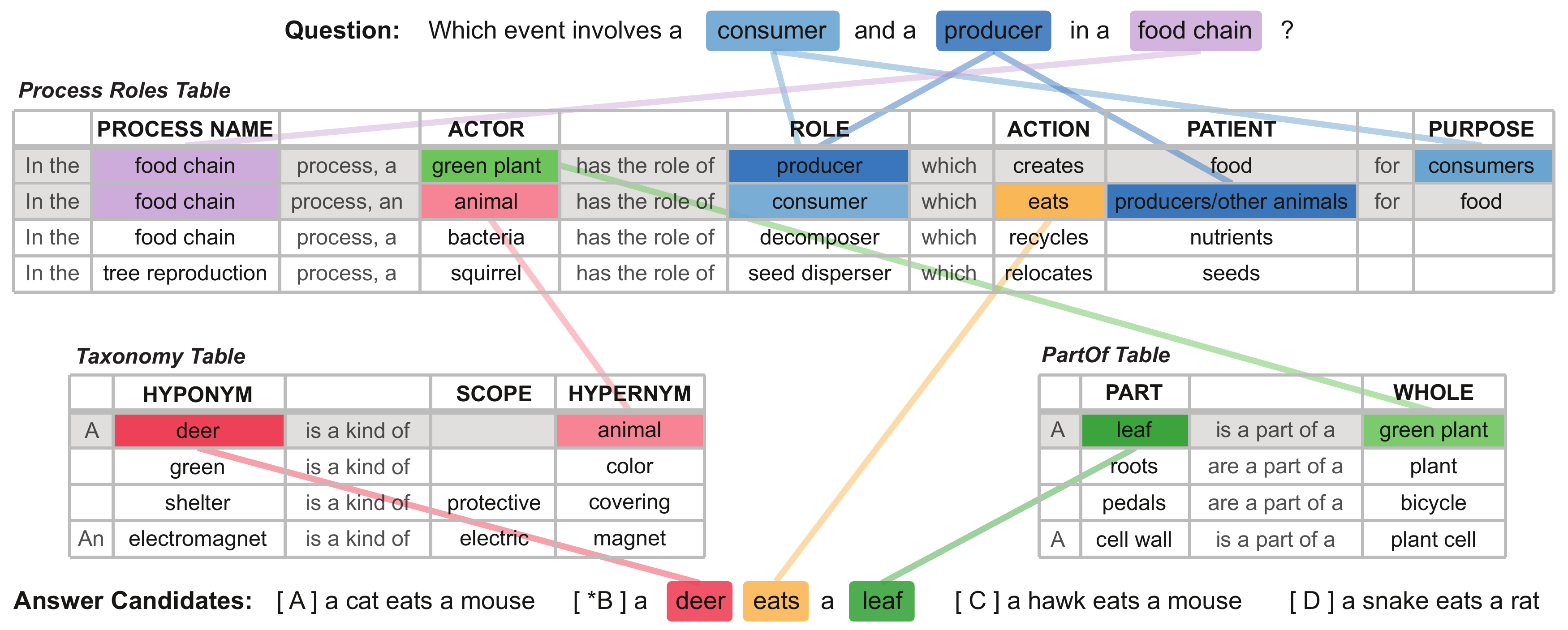} 
\vspace{-4mm}
\caption[]{\small Examples of tables and table rows from the tablestore, grounded in an example question and explanation.  Table columns define the primary roles or arguments for a given relation (e.g. \textit{process name, actor, role, etc}).  Unlabeled ``filler'' columns allow each row to be used as a stand-alone natural language sentence.  Note that for clarity only 4 example rows per table are shown.\footnotemark }
\label{fig:tablestore}
\end{center}
\vspace{-4mm}
\end{figure*}

\noindent Explanations for a given question consist of a set of sentences, each of which is on a single topic and centered around a particular kind of relation, such as \textit{water is a kind of liquid (a taxonomic relation)}, or \textit{melting means changing from a solid to a liquid through the addition of heat energy (a change relation)}.  

Each explanation sentence is represented as a single row from a semi-structured table defined around a particular relation.  Our tablestore includes 62 such tables, each centered around a particular relation such as \textit{taxonomy, meronymy, causality, changes, actions, requirements,} or \textit{affordances}, and a number of tables specified around specific properties, such as \textit{average lifespans of living things, the magnetic properties of materials,} or the \textit{nominal durations of certain processes (like the Earth orbiting the Sun)}.  The initial selection of table relations was drawn from a list of 21 common relations required for science explanations identified by Jansen et al. \shortcite{jansen2016:COLING} on a smaller corpus, and expanded as new knowledge types were identified.  Subsets of example tables are included in Figure \ref{fig:tablestore}.  Each explanation in this corpus contains an average of 6.3 rows. 

{\flushleft \textbf{Fine-grained column structure:}} In tabular representations, columns represent specific roles or arguments to a specific relation (such as \textit{X is when Y changes from A to B using mechanism C}).  In our tablestore we attempt to minimize the amount of information per cell, instead favouring tables with many columns that explicitly identify common roles, conditions, or other relations.  This finer-grained structure eases the annotator's cognitive load when authoring new rows, while also better compartmentalizing the relational knowledge in each row for inference algorithms.  The tables in our tablestore contain between \textit{2 and 16} content columns, as compared to 2 to 5 columns for the Ariso tablestore \cite{Khashabi:2016TableILP}.


{\flushleft \textbf{Natural language sentences:}} QA models use a variety of different representations for inference, from semantic roles and syntactic dependencies to discourse and embeddings.  Following Khashabi et al. \shortcite{Khashabi:2016TableILP}, we make use of a specific form of table representation that includes ``filler'' columns that allow each row to be directly read off as a stand-alone natural language sentence, and serve as input to any model.  Examples of these filler columns can be seen in Figure \ref{fig:tablestore}.

\subsection{Explanation Graphs and Sentence Roles}

%
%
\begin{table}[t]
\footnotesize
\centering
\begin{tabular}{lll}
Question	&	\multicolumn{2}{l}{ Which occurs as the kinetic energy of water}\\
			&	\multicolumn{2}{l}{ molecules increases? }\\
Answer		&	[*D] liquid water becomes water vapor	\\

\hline

\multicolumn{3}{l}{ \textit{Central role} }\\
\multicolumn{3}{l}{ ~~As a molecule's kinetic energy increases, temperature will }\\
\multicolumn{3}{l}{ ~~ increase. }\\
\multicolumn{3}{l}{ ~~Boiling means changing from a liquid into a gas by adding }\\
\multicolumn{3}{l}{ ~~ heat energy. }\\
\\
\multicolumn{3}{l}{ \textit{Grounding role} }\\
\multicolumn{3}{l}{ ~~Water is a kind of liquid. }\\
\multicolumn{3}{l}{ ~~Water is in the gas state, called water vapor, for }\\
\multicolumn{3}{l}{ ~~ temperatures greater than 100 degrees celsius.}\\
\\
\multicolumn{3}{l}{ \textit{Background role} }\\
\multicolumn{3}{l}{ ~~Matter is made of molecules. }\\
\\
\multicolumn{3}{l}{ \textit{Lexical glue role} }\\
\multicolumn{3}{l}{ ~~To add means to increase. }\\
\multicolumn{3}{l}{ ~~Temperature is a measure of heat energy. }\\
\hline

\end{tabular}
\caption{\small Examples of the four coarse classes of explanation sentence roles, \textit{central, grounding, background, and lexical glue}. }
\label{tab:explanationroles}
\vspace{-2mm}
\end{table}

\noindent Explanations for a given question here take the form of a list of sentences, where each sentence is a reference to a specific table row in the table store.  To increase their utility for knowledge and inference analyses, we require that each sentence in an explanation be explicitly lexically connected (i.e. share words) with either the question, answer, or other sentences in the explanation.  We call this lexically-connected set of sentences an \textit{explanation graph}. 

In our preliminary analysis, we observed that the sentences in our explanations can take on very different roles, and we hypothesize that differentiating these roles is likely important for inference algorithms.  We identified four coarse roles, listed in Table \ref{tab:explanationroles}, and described below:

\begin{itemize}
\item \textbf{Central:} The central concept(s) that a question is testing, such as \textit{changes of state} or the \textit{coupled relationship between kinetic energy and temperature}. 


\item \textbf{Grounding:} Sentences linking generic or abstract terms in a \textit{central} sentence with specific instances of those terms in the question or answer.  For example, for questions about changes of state, grounding sentences might identify specific instances of \textit{liquids} (such as water) or \textit{gasses} (such as water vapor). 

\item \textbf{Background:} Extra information elaborating on the topic, but that (strictly speaking) isn't required to arrive at the correct inference. 

\item \textbf{Lexical glue:} Sentences that lexically link two concepts, such as \textit{``to add means to increase''}, or \textit{``heating means adding heat''}.  This is an artificial category in our corpus, brought about by the need for explanation graphs to be explicitly lexically linked. 

\end{itemize}

\noindent For each sentence in each authored explanation, we provide annotation indicating which of these four roles the sentence serves in that explanation. 

\footnotetext{Note that this figure also appears in an earlier workshop submission on identifying explanatory patterns \cite{jansen:akbc2017}}

\subsection{Annotation Tool}

\begin{figure}[t]
\begin{center}
\includegraphics[scale=0.145]{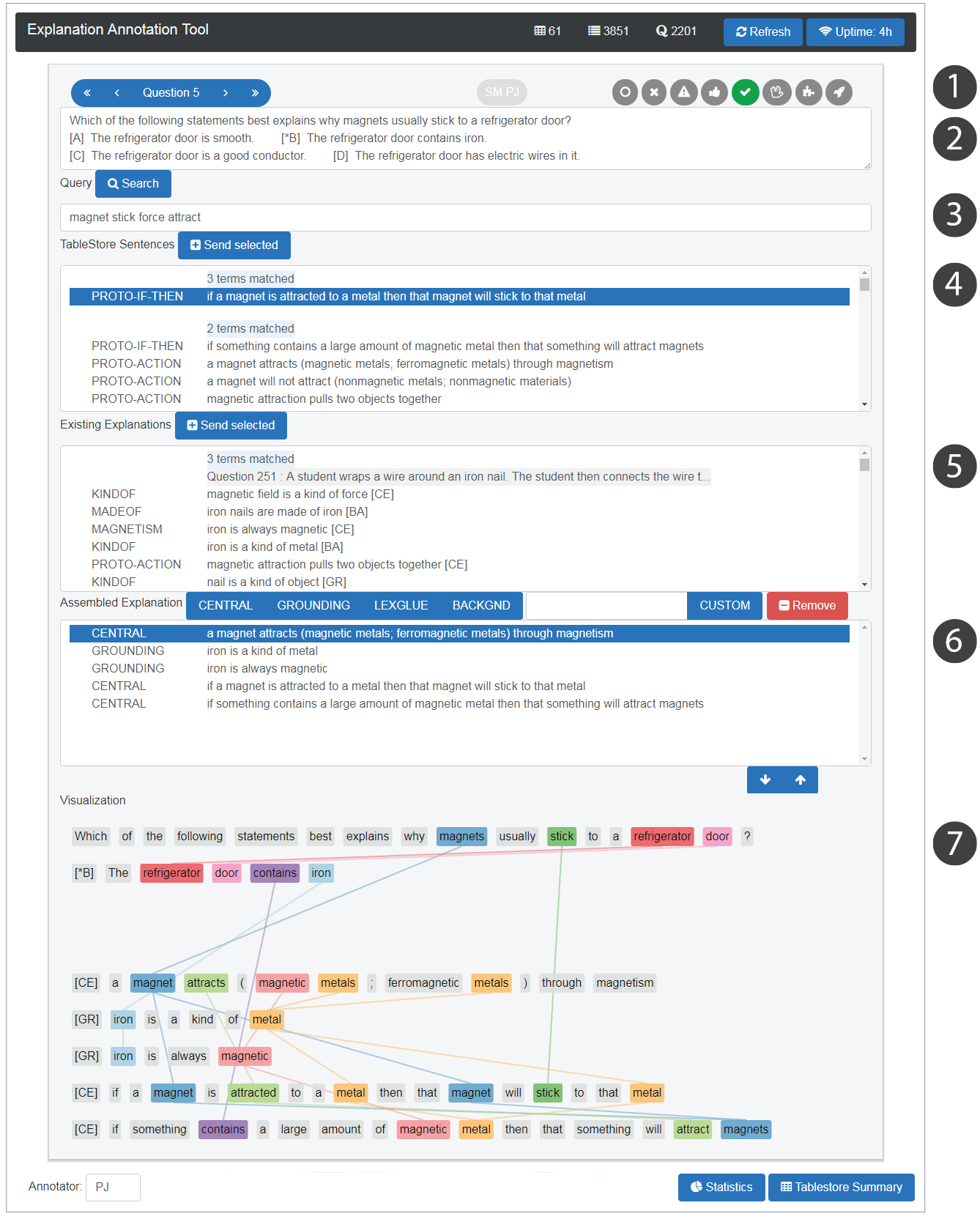} 
\caption{The explanation authoring web tool.  Interface components include: (1) A list of user-settable flags to assist in the annotation and quality review process; (2) Question and answer candidates; (3) Query terms for search; (4) Query results (tablestore); (5) Query results (complete explanations); (6) Current explanation being assembled; (7) Explanation graph visualization of lexical overlap within the explanation.  }
\label{fig:tool}
\end{center}
\vspace{-4mm}
\end{figure}

\noindent To facilitate explanation authoring, we developed and iterated the web-based collaborative authoring tool shown in Figure \ref{fig:tool}.  The tool displays a given question to the explanation author, and allows the author to progressively build an explanation graph for that question by querying the tablestore for relevant rows based on keyword searches, as well as past explanations that are likely to contain similar content or structure (increasing consistency across explanations, while reducing annotation time).  A graphical visualization of the explanation graph helps the author quickly assess gaps in the explanation content to address by highlighting lexical overlap between sentences with coloured edges and labels. The tablestore takes the form of a shared Google Sheet\footnote{\url{http://sheets.google.com}} that the annotators populate, with each table represented as a separate tab on the sheet.

%



\subsection{Procedure and Explanation Review}
\noindent For a given question, annotators identified the central concept the question was testing, as well as the inference required to correctly answer the question, then began progressively constructing the explanation graph.  Sentences in the graph were added by querying the tablestore based on keywords, which retrieved both single sentences/table rows, as well as entire explanations that had been previously annotated.  If any knowledge required to build an explanation did not exist in the tablestore, this was added to an appropriate table, then added to the explanation. 



New tables were regularly added, most commonly for property knowledge surrounding a particular topic (e.g. whether a particular material is recyclable).  Because explanations are stored as lists of unique identifiers to table rows, tables and table rows could regularly be refactored, elaborated, or entirely reorganized without requiring existing explanations to be rewritten.  We found this was critical for consistency and ensuring good organization throughout corpus construction.

One of the central difficulties with evaluating explanation authoring is determining metrics for interannotator agreement, as many correct explanations are possible for a given question, and there are many different wordings that an annotator might choose to express a given piece of knowledge in the tablestore.  Similarly, the borders between different levels of explanatory depth are fuzzy, suggesting that one annotator may express their explanation with more or less specificity than another. 

To address these difficulties we included two methods to increase consistency.  First, as a passive intervention during the explanation generation process, annotators are presented with existing explanations that can be drawn from to compose a new explanation, where these existing explanations share many of the same query terms being used to construct the new explanation.  Second, as an active intervention, each explanation goes through four review passes to ensure consistency.  The first two passes are completed by the original annotator, before checking a flag on the annotation tool signifying that the question is ready for external review.  A second annotator then checks the question for completeness and consistency with existing explanations, and composes a list of suggested edits and revisions.  The fourth and final pass is completed by the original annotator, who implements these suggested revisions. 
This review process is expensive, taking approximately one third of the total time required to annotate each question. 

Each annotator required approximately 60 hours of initial training for this explanation authoring task.  We found that most explanations could be constructed within 5-10 minutes, with the review process taking approximately 5 more minutes per question.

\section{Explanation Corpus Properties}

\noindent Here we characterize three properties of the explanation corpus as they relate to developing methods of explainable inference: \textit{ knowledge frequency, explanation overlap,} and \textit{tablestore growth.}

\subsection{Knowledge Use and Row Frequency}


\noindent The tables most frequently used to author explanations are shown in Table \ref{tab:knowledgeuse}, broken down into three broad categories identified by Jansen et al. \shortcite{jansen2016:COLING}: \textit{retrieval types, inference-supporting types, and complex inference types}.  Because the design of this corpus is data driven -- i.e., knowledge is generally added to a table because it is required in one or more explanations\footnote{For compatibility, we do include several property tables from the Aristo tablestore, though a large proportion of rows from these tables are not actively used.  Our tablestore includes 4,950 rows, 3,686 of which are actively used in at least one explanation.} -- we can calculate how frequently the rows in a given table are reused to obtain an approximate measure of the generality of that knowledge.  On average, a given table row is used in 2.9 different explanations, with 1,535 rows used more than once, and 531 rows used 5 or more times. The most frequently reused row (\textit{''an animal is a kind of organism''}) is used in 89 different explanations.  Generic ``change of state'' knowledge (e.g. solids, liquids, and gasses) is also frequently reused, with each row in the StatesOfMatter table used in an average of 15.7 explanations.  Usage statistics for other common tables are also provided in Table \ref{tab:knowledgeuse}.

%
%
\begin{table}[t]
\footnotesize
\centering
\begin{tabular}{lccc}

			 	&	Prevalence		&	Rows in		&	Avg. Row	\\
Knowledge Type	&	(\% of expl.)	&	Table 		&	Freq.	\\
\hline

\multicolumn{4}{l}{\textit{Retrieval Types}}\\
~~Taxonomic				&	78\%	&	1,119		&	1.2 \\
~~Synonymy				&	61\%	&	639			&	1.6 \\
~~PartOf				&	14\%	&	148			&	1.6\\
~~Properties (Generic)	&	11\%	&	173			&	1.1\\
~~MadeOf				&	7\%		&	72			&	1.7\\
~~Contains				&	6\%		&	75			&	1.4\\
~~Examples				&	5\%		&	58			&	1.4\\
~~Measurements (P)		&	4\%		&	23			&	3.0\\
~~Locations	(P)			&	3\%		&	47			&	1.1\\
~~InheritedTraits (P)	&	3\%		&	22			&	2.3\\
~~StatesOfMatter (P)	&	3\%		&	3			&	15.7\\
~~Conductivity (P)		&	3\%		&	9			&	4.9\\
~~Resources (P)			&	3\%		&	16			&	2.7\\

\\

\multicolumn{4}{l}{\textit{Inference Supporting Types}}\\
~~Actions				&	25\%	&	259			&	1.6 \\
~~UsedFor				&	19\%	&	191			&	1.7 \\
~~Requires				&	15\%	&	121			&	2.1\\
~~SourceOf				&	14\%	&	81			&	2.8\\
~~Affect				&	12\%	&	77			&	2.6\\
~~Opposites				&	8\%		&	35			&	3.8\\
~~FormedBy				&	4\%		&	40			&	1.9\\
~~Affordances			&	4\%		&	48			&	1.3\\
\\

\multicolumn{4}{l}{\textit{Complex Inference Types}}\\
~~If/Then				&	21\%	&	229			&	1.6 \\
~~Cause					&	17\%	&	183			&	1.6 \\
~~Changes (discrete)	&	14\%	&	62			&	3.8\\
~~Transfer				&	9\%		&	46			&	3.3\\
~~Changes (vector)		&	9\%		&	62			&	2.4\\
~~CoupledRelationships	&	7\%		&	126			&	0.9\\
~~ProcessRoles			&	3\%		&	12			&	3.8\\

\hline

\end{tabular}
\caption{\small The proportion of explanations that contain knowledge from a given table, sorted by most frequent knowledge, and broken down by the knowledge type of a given table.  Tables not used in at least 3\% of explanations are not shown. (P) indicates a given table describes properties, e.g. whether a given material is conductive. Average Row Frequency refers to the average number of explanations a given row from that table is used in. }
\label{tab:knowledgeuse}
\vspace{-4mm}
\end{table}

\subsection{Explanation Overlap}

\noindent One might hypothesize that questions that require similar inferences to correctly answer may also contain some of the same knowledge in their explanations, with the amount of knowledge overlap dependent upon the similarity of the questions.  We plan to explore using this overlap as a method of inference that can generate new explanations by editing, merging, or expanding known explanations from similar, known questions (see Jansen \shortcite{jansen:akbc2017} for an initial study).  For this to be possible, an explanation corpus must reach a sufficient size that a large majority of questions have substantial overlap in their explanations. 

\begin{figure*}[!h]
\begin{center}
\includegraphics[scale=0.30]{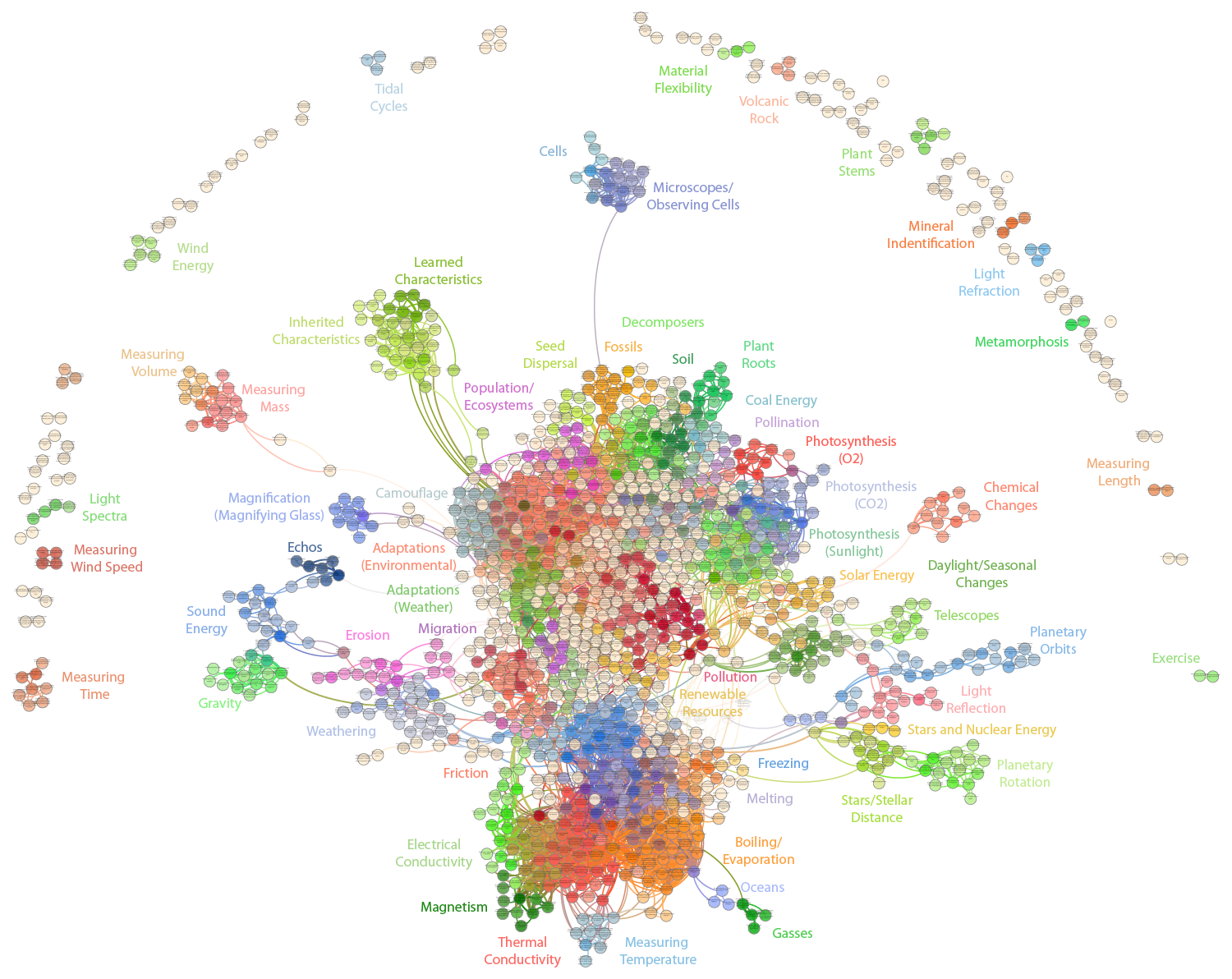} 
\caption{\small Questions in this explanation corpus connected by explanation overlap.  Here, nodes represent questions and their explanations, and edges between nodes represent two questions having \textit{at least 2 or more (i.e. 2+)} shared rows (i.e. sentences) in their explanations, with at least one of these shared rows being labelled as having a CENTRAL role to the explanation.  Topic clusters (labels) naturally emerge for questions requiring similar methods of inference, based on the shared content of their explanations. }
\label{fig.worldtree}
\end{center}
\vspace{-2mm}
\end{figure*}

\begin{figure}[!t]
\begin{center}
\includegraphics[scale=0.30]{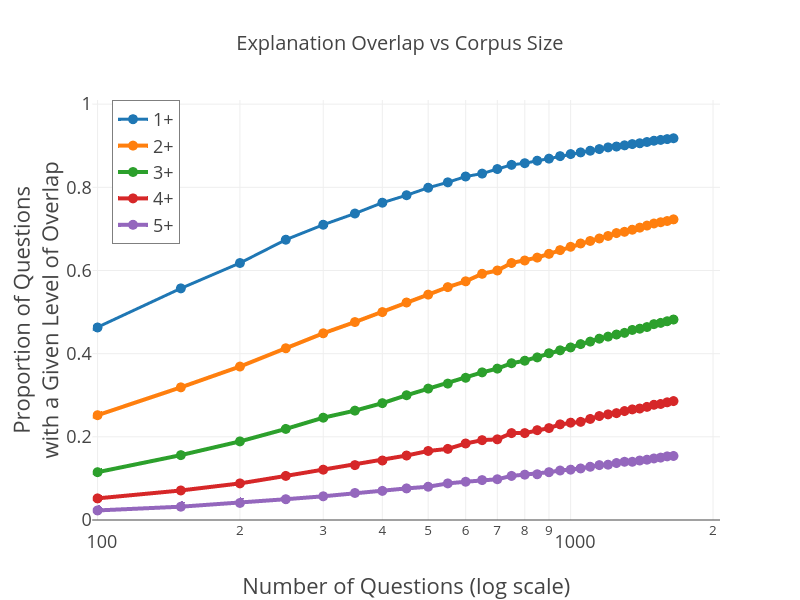} 
\caption{\small Monte-carlo simulation showing the proportion of questions whose explanations overlap by \textit{1 or more, 2 or more, 3 or more, ...,} explanation sentences.  The proportion increases logarithmically with the number of questions in the corpus. Each point represents the average of 100 simulations.} 
\label{fig.explanationoverlap}
\end{center}
\vspace{-4mm}
\end{figure}

Figure \ref{fig.explanationoverlap} shows the proportion of questions in the corpus that have \textit{1 or more}, \textit{2 or more}, \textit{3 or more}, \textit{etc.}, overlapping rows in their explanations with \textit{at least one other question} in the corpus.\footnote{Though not included for space, the number of questions with \textit{N or more} rows in common in their explanations increases linearly with the number of questions.  For this corpus, for a given question, on average there are 17 questions that have 1 or more overlapping rows in their explanation, 9 questions with 2 or more shared rows in their explanation, and 5 questions with 3 or more shared rows in their explanation.}  Similarly, to ground this, Figure \ref{fig.worldtree} shows a visualization of questions whose explanations have \textit{2 or more} overlapping rows.  For a given level of overlapping explanation sentences, Figure \ref{fig.explanationoverlap} shows that the proportion of questions with that level of overlap increases logarithmically with the number of questions.

This has two consequences.  First, it allows us to estimate the size of corpus required to train hypothetical inference methods for the science exam domain capable of producing explanations.  If a given inference method can work successfully with only minimal overlap (for example, 1 shared table row), then a training corpus of 500 explanations in this domain should be sufficient to answer 80\% of questions.  If an inference method requires 2 shared rows, the corpus requirements would increase to approximately 2,500 questions to answer 80\% of questions.  However, if an inference method requires 3 or more rows, this likely would not be possible without a corpus of at least 20,000 questions and explanations -- a substantial undertaking.  Second, because this relationship is strongly logarithmic, if it transfers to domains outside elementary science, it should be possible to estimate the corpus size requirements for those domains after authoring explanations for only a few hundred questions.

\subsection{Explanation Tablestore Growth}

\begin{figure}[!t]
\begin{center}
\includegraphics[scale=0.30]{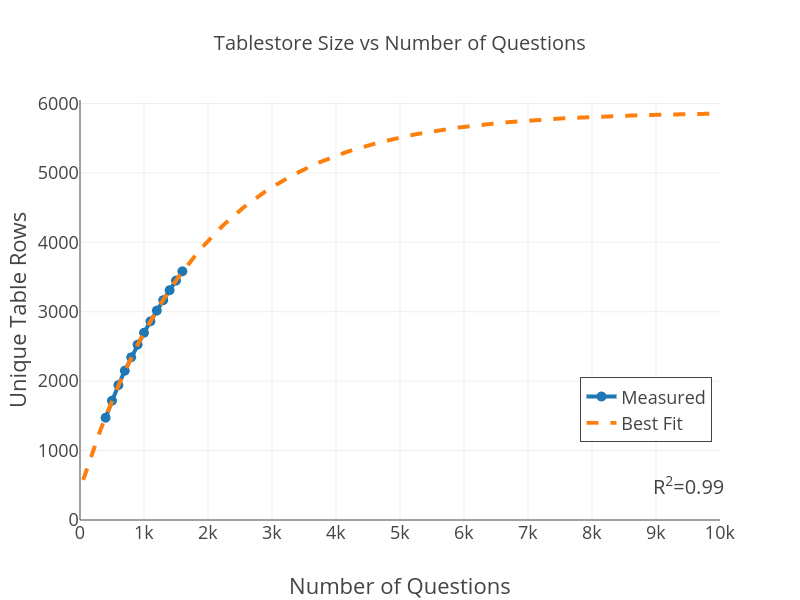} 
\caption{\small Monte-carlo simulation showing the number of unique table rows required to explainably answer a given number of questions.  The line of best fit (dashed) suggests that this is a proportional decay relationship ($R^{2}=0.99$), asymptoting at approximately 6,000 table rows and 10,000 questions. Each point represents the average of 10,000 simulations. }
\label{fig.tablestoregrowth}
\end{center}
\vspace{-4mm}
\end{figure}

Finally, we examine the growth of the tablestore as it relates to the number of questions in the corpus. Figure \ref{fig.tablestoregrowth} shows a monte-carlo simulation of the number of unique tablestore rows required to author explanations for specific corpus sizes.  This relationship is strongly correlated \textit{(R=0.99)} with an exponential proportional decrease.\footnote{Here, this exponential proportional decrease takes the form of $R = 434 - (-2.93/0.00054) \cdot (1-e^{-0.00054 \cdot Q}),$ where $R$ is the size of the tablestore in rows, to explainably answer $Q$ questions.}  For this elementary science corpus, this asymptotes at approximately 6,000 unique table rows, and 10,000 questions, providing an estimate of the upper-bound of knowledge required in this domain, and the number of unique questions that can be generated within the scope of the elementary science curriculum. 

The caveat to this estimate is that it estimates the knowledge required for elementary science exams as they currently exist, with the natural level of variation introduced by the test designers.  Questions are naturally grounded in examples, such as \textit{``Which part of an oak tree is responsible for undertaking photosynthesis?'' (Answer: the leaves)}.  While the corpus often contains a number of variations of a given question that test the same curriculum topic and have similar explanations, many more variations on these questions are possible that ground the question in different examples, like \textit{orchids}, \textit{peach trees}, or other plants.  As such, while we believe that these estimates likely cover the core knowledge of the domain, many times that knowledge would be required to make the explanation tablestore robust to small variations in the presentation of those existing exam questions, or to novel unseen questions.

%
%
%
%
%
%
%
%
%
%
%
%
%
%
%
%

\section{Conclusion}

\noindent We provide a corpus of explanation graphs for elementary science questions suitable for work in developing explainable methods of inference, and show that the knowledge frequency, explanation overlap, and tablestore growth properties of the corpus follow predictable relationships.  This work is open source, with the corpus and generation tools available at \url{http://www.cognitiveai.org/explanationbank}.

\section{Acknowledgements}

\noindent We thank the Allen Institute of Artificial Intelligence for funding this work, Peter Clark at AI2 for thoughtful discussions, and Paul Hein for assistance constructing the annotation tool.

\section{Bibliographical References}
\label{main:ref}

\bibliographystyle{lrec2016}
\bibliography{refs}

\begin{thebibliography}{}

\bibitem[\protect\citename{{Speecon Consortium}}2014]{speecon}
{Speecon Consortium}.
\newblock (2014).
\newblock {\em Dutch Speecon Database}.
\newblock Speecon Project, distributed via ELRA, Speecon resources, 1.0, ISLRN
  613-489-674-355-0.

\end{thebibliography}


\begin{thebibliography}{}

\bibitem[\protect\citename{Berant \bgroup et al.\egroup
  }2013]{berant2013semantic}
Berant, J., Chou, A., Frostig, R., and Liang, P.
\newblock (2013).
\newblock Semantic parsing on freebase from question-answer pairs.
\newblock In {\em EMNLP}.

\bibitem[\protect\citename{Clark \bgroup et al.\egroup }2013]{clark:2013}
Clark, P., Harrison, P., and Balasubramanian, N.
\newblock (2013).
\newblock A study of the knowledge base requirements for passing an elementary
  science test.
\newblock In {\em Proceedings of the 2013 Workshop on Automated Knowledge Base
  Construction}, AKBC'13, pages 37--42.

\bibitem[\protect\citename{Clark \bgroup et al.\egroup
  }2016]{Clark2016CombiningRS}
Clark, P., Etzioni, O., Khot, T., Sabharwal, A., Tafjord, O., Turney, P.~D.,
  and Khashabi, D.
\newblock (2016).
\newblock Combining retrieval, statistics, and inference to answer elementary
  science questions.
\newblock In {\em Proceedings of the Thirtieth {AAAI} Conference on Artificial
  Intelligence, February 12-17, 2016, Phoenix, Arizona, {USA.}}, pages
  2580--2586.

\bibitem[\protect\citename{Clark}2015]{clark:2015}
Clark, P.
\newblock (2015).
\newblock Elementary school science and math tests as a driver for {AI:} take
  the aristo challenge!
\newblock In Blai Bonet et~al., editors, {\em Proceedings of the Twenty-Ninth
  {AAAI} Conference on Artificial Intelligence, January 25-30, 2015, Austin,
  Texas, {USA.}}, pages 4019--4021. {AAAI} Press.

\bibitem[\protect\citename{Dalvi \bgroup et al.\egroup }2016]{Dalvi2016IKE}
Dalvi, B., Bhakthavatsalam, S., Clark, C., Clark, P., Etzioni, O., Fader, A.,
  and Groeneveld, D.
\newblock (2016).
\newblock {IKE} - an interactive tool for knowledge extraction.
\newblock In {\em Proceedings of the 5th Workshop on Automated Knowledge Base
  Construction, AKBC@NAACL-HLT 2016, San Diego, CA, USA, June 17, 2016}, pages
  12--17.

\bibitem[\protect\citename{Etzioni \bgroup et al.\egroup
  }2011]{etzioni:2011open}
Etzioni, O., Fader, A., Christensen, J., Soderland, S., and Mausam, M.
\newblock (2011).
\newblock Open information extraction: The second generation.
\newblock In {\em Proceedings of the 20th International Joint Conference on
  Artificial Intelligence (ICJAI)}, pages 3--10.

\bibitem[\protect\citename{Fried \bgroup et al.\egroup }2015]{fried2015higher}
Fried, D., Jansen, P., Hahn-Powell, G., Surdeanu, M., and Clark, P.
\newblock (2015).
\newblock Higher-order lexical semantic models for non-factoid answer
  reranking.
\newblock {\em Transactions of the Association for Computational Linguistics},
  3:197--210.

\bibitem[\protect\citename{Jansen \bgroup et al.\egroup
  }2016]{jansen2016:COLING}
Jansen, P., Balasubramanian, N., Surdeanu, M., and Clark, P.
\newblock (2016).
\newblock What's in an explanation? characterizing knowledge and inference
  requirements for elementary science exams.
\newblock In {\em Proceedings of COLING 2016, the 26th International Conference
  on Computational Linguistics: Technical Papers}, pages 2956--2965, Osaka,
  Japan, December. The COLING 2016 Organizing Committee.

\bibitem[\protect\citename{Jansen \bgroup et al.\egroup
  }2017]{jansen2017framing}
Jansen, P., Sharp, R., Surdeanu, M., and Clark, P.
\newblock (2017).
\newblock Framing qa as building and ranking intersentence answer
  justifications.
\newblock {\em Computational Linguistics}.

\bibitem[\protect\citename{Jansen}2017]{jansen:akbc2017}
Jansen, P.
\newblock (2017).
\newblock A study of automatically acquiring explanatory inference patterns
  from corpora of explanations: Lessons from elementary science exams.
\newblock In {\em Proceedings of the 2017 Workshop on Automated Knowledge Base
  Construction}, AKBC'17.

\bibitem[\protect\citename{Jauhar \bgroup et al.\egroup
  }2016]{jauhar:2016tables}
Jauhar, S.~K., Turney, P.~D., and Hovy, E.~H.
\newblock (2016).
\newblock Tables as semi-structured knowledge for question answering.
\newblock In {\em Proceedings of the 54th Annual Meeting of the Association for
  Computational Linguistics (ACL)}.

\bibitem[\protect\citename{Khashabi \bgroup et al.\egroup
  }2016]{Khashabi:2016TableILP}
Khashabi, D., Khot, T., Sabharwal, A., Clark, P., Etzioni, O., and Roth, D.
\newblock (2016).
\newblock Question answering via integer programming over semi-structured
  knowledge.
\newblock In {\em Proceedings of the International Joint Conference on
  Artificial Intelligence}, IJCAI'16, pages 1145--1152.

\bibitem[\protect\citename{Khot \bgroup et al.\egroup
  }2015]{Khot2015ExploringML}
Khot, T., Balasubramanian, N., Gribkoff, E., Sabharwal, A., Clark, P., and
  Etzioni, O.
\newblock (2015).
\newblock Exploring markov logic networks for question answering.
\newblock In {\em EMNLP}.

\bibitem[\protect\citename{Khot \bgroup et al.\egroup }2017]{Khot:ACL2017}
Khot, T., Sabharwal, A., and Clark, P.
\newblock (2017).
\newblock Answering complex questions using open information extraction.
\newblock In {\em Proceedings of the 55th Annual Meeting of the Association for
  Computational Linguistics, {ACL} 2017, Vancouver, Canada, July 30 - August 4,
  Volume 2: Short Papers}, pages 311--316.

\bibitem[\protect\citename{Legare}2014]{legare2014contributions}
Legare, C.~H.
\newblock (2014).
\newblock The contributions of explanation and exploration to children's
  scientific reasoning.
\newblock {\em Child Development Perspectives}, 8(2):101--106.

\bibitem[\protect\citename{Pasupat and Liang}2015]{pasupat:2015}
Pasupat, P. and Liang, P.
\newblock (2015).
\newblock Compositional semantic parsing on semi-structured tables.
\newblock In {\em Proceedings of the 53rd Annual Meeting of the Association for
  Computational Linguistics (ACL)}.

\bibitem[\protect\citename{Ribeiro \bgroup et al.\egroup }2016]{Ribeiro2016}
Ribeiro, M.~T., Singh, S., and Guestrin, C.
\newblock (2016).
\newblock "why should i trust you?": Explaining the predictions of any
  classifier.
\newblock In {\em Proceedings of the 22Nd ACM SIGKDD International Conference
  on Knowledge Discovery and Data Mining}, KDD '16, pages 1135--1144, New York,
  NY, USA. ACM.

\bibitem[\protect\citename{Rittle-Johnson and Loehr}2016]{rittle2016eliciting}
Rittle-Johnson, B. and Loehr, A.~M.
\newblock (2016).
\newblock Eliciting explanations: Constraints on when self-explanation aids
  learning.
\newblock {\em Psychonomic bulletin \& review}, pages 1--10.

\bibitem[\protect\citename{Roscoe and Chi}2007]{roscoe2007understanding}
Roscoe, R.~D. and Chi, M.~T.
\newblock (2007).
\newblock Understanding tutor learning: Knowledge-building and
  knowledge-telling in peer tutors' explanations and questions.
\newblock {\em Review of Educational Research}, 77(4):534--574.

\bibitem[\protect\citename{Schoenick \bgroup et al.\egroup
  }2017]{schoenick2017moving}
Schoenick, C., Clark, P., Tafjord, O., Turney, P., and Etzioni, O.
\newblock (2017).
\newblock Moving beyond the turing test with the allen ai science challenge.
\newblock {\em Communications of the ACM}, 60(9):60--64.

\bibitem[\protect\citename{Sharp \bgroup et al.\egroup
  }2017]{sharp:2017tellmewhy}
Sharp, R., Surdeanu, M., Jansen, P., Valenzuela-Escárcega, M.~A., Clark, P.,
  and Hammond, M.
\newblock (2017).
\newblock Tell me why: Using question answering as distant supervision for
  answer justification.
\newblock In {\em Proceedings of the Conference on Computational Natural
  Language Learning (CoNLL)}.

\bibitem[\protect\citename{Sun \bgroup et al.\egroup }2016]{sun:2016table}
Sun, H., Ma, H., He, X., Yih, W.-t., Su, Y., and Yan, X.
\newblock (2016).
\newblock Table cell search for question answering.
\newblock In {\em Proceedings of the 25th International Conference on World
  Wide Web (WWW)}, pages 771--782.

\bibitem[\protect\citename{Yin \bgroup et al.\egroup }2015]{yin:2015answering}
Yin, P., Duan, N., Kao, B., Bao, J., and Zhou, M.
\newblock (2015).
\newblock Answering questions with complex semantic constraints on open
  knowledge bases.
\newblock In {\em Proceedings of the 24th ACM International on Conference on
  Information and Knowledge Management}, pages 1301--1310. ACM.

\end{thebibliography}


\end{document}